\documentclass{article}

     \usepackage[preprint, nonatbib]{neurips_2020}

\usepackage[utf8]{inputenc} 
\usepackage[T1]{fontenc}    
\usepackage{url}            
\usepackage{booktabs}       
\usepackage{amsfonts}       
\usepackage{nicefrac}       
\usepackage{microtype}      
\usepackage{xcolor}         
\usepackage{amsmath}
\usepackage{graphicx}
\usepackage{subfig}
\usepackage{float}
\usepackage{tcolorbox}
\usepackage[style=numeric, sorting=none]{biblatex}
\usepackage{amsthm}
\usepackage{bm}
\usepackage{stmaryrd}
\usepackage{wrapfig}
\usepackage{hyperref}
\usepackage{mathtools}
\usepackage{algorithm} 
\usepackage{algorithmic}  
\usepackage[algo2e]{algorithm2e} 

\title{Self-organization of multi-layer spiking neural networks}

\author{%
  Guruprasad Raghavan$^*$\\
  Caltech\\
  Pasadena, CA 91106 \\
  \texttt{graghava@caltech.edu} \\
  \And
  Cong Lin$^*$\\
  Caltech \\
  Pasadena, CA 91106 \\ 
  \texttt{conglin@caltech.edu}
  \And
  Matt Thomson\\
  Caltech\\
  Pasadena, CA 91106 \\
  \texttt{mthomson@caltech.edu} \\
}

\begin{document}

\maketitle

\begin{abstract}
Living neural networks in our brains autonomously self-organize into large, complex architectures during early development to result in an organized and functional organic computational device. A key mechanism that enables the formation of complex architecture in the developing brain is the emergence of traveling spatio-temporal waves of neuronal activity across the growing brain. Inspired by this strategy, we attempt to efficiently self-organize large neural networks with an arbitrary number of layers into a wide variety of architectures. To achieve this, we propose a modular tool-kit in the form of a dynamical system that can be seamlessly stacked to assemble multi-layer neural networks. The dynamical system encapsulates the dynamics of spiking units, their inter$\addslash$intra layer interactions as well as the plasticity rules that control the flow of information between layers. The key features of our tool-kit are (1) autonomous spatio-temporal waves across multiple layers triggered by activity in the preceding layer and (2) Spike-timing dependent plasticity (STDP) learning rules that update the inter-layer connectivity based on wave activity in the connecting layers. Our framework leads to the self-organization of a wide variety of architectures, ranging from multi-layer perceptrons to autoencoders. We also demonstrate that emergent waves can self-organize spiking network architecture to perform unsupervised learning, and networks can be coupled with a linear classifier to perform classification on classic image datasets like MNIST. Broadly, our work shows that a dynamical systems framework for learning can be used to self-organize large computational devices.

\end{abstract}

\section{Introduction}

Biological neural networks in brains are remarkable machines that endow an organism with the ability to perform an array of computational and information processing tasks  \cite{glickfeld2017higher,peirce2015understanding,tan2017development,denny2017engrams,hanks2017perceptual, padoa2017orbitofrontal}. In addition, biological neural networks are fascinating as they grow from a single precursor cell and self-organize into complex architectures \cite{singer1986brain, singer2009brain}. The self-organization process in biological networks leads to a wide variety of architectures ranging from feed-forward networks for visual processing in the visual cortex \cite{lewis2005self} to recurrent neural networks for memory systems deployed in the hippocampus \cite{buzsaki2005synaptic}.

One of the key mechanisms that guides the self-organization process in a developing embryo's neural networks is the emergence of spatio-temporal neural activity waves across multiple regions of the brain \cite{penn1999brain, oldham2019development, godfrey2007retinal}. Traveling activity waves in the developing brain carry significant information to achieve two major purposes: (i) wiring local networks into specific architectures and (ii) for initiating the maturation of neural circuitry \cite{donato2017stellate}. 

The first demonstration of utilizing spontaneous traveling waves to self-organize a two-layered neural network was shown in \cite{raghavan2019neural}. Although successful in self-organizing retinotopic pooling layers of variable pool-sizes, the strategy was limited to a two layered neural network. Neural networks composed of spiking nodes are of great interest to the fields of AI and neuroscience, for they model the dynamics of neurons in our brains closely, can be trained to perform AI-relevant tasks through strategies that are more biologically plausible, are apt models to study self-organization of living neural systems and can be implemented on neuromorphic hardware. 

In this paper, we develop strategies to self-organize large spatially-connected, multi-layer spiking neural networks (SNN), inspired by the wiring rules and mechanisms adopted by the mammalian visual system during development. The visual circuitry, specifically the connectivity between the retina, LGN and the early layers of the visual cortex have stereotypical architectures across organisms, namely pooling connectivity between retina and LGN, and an expansion from the LGN to V1 \cite{coen2017method}. The connectivity is established by the emergence of multiple traveling waves (figure-1) across the retina and different cortical regions much before the onset of vision. 

\begin{wrapfigure}{r}{0.5\textwidth}
  \begin{center}
    \includegraphics[scale=0.4]{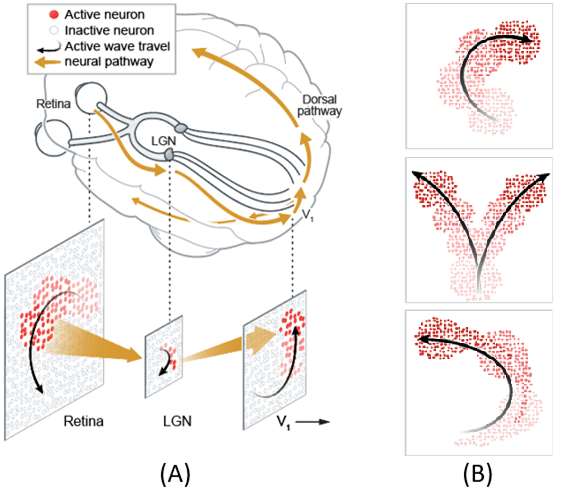}
  \end{center}
  \caption{\textbf{Spontaneous waves in the developing brain} \textbf{(A)} Emergent neuronal waves across the visual circuitry (Retina, LGN and V1). \textbf{(B)} Multiple types of wave dynamics.}
  \label{fig:bioInspired}
\end{wrapfigure}


The main contribution of this paper is that we propose a modular tool-kit in the form of a dynamical systems framework to seamlessly self-organize large neural networks, inspired by cortical developmental processes. The modular structure of the tool-kit allows us to scale the network on demand and rapidly evolve neural architectures, by modifying the components of a module. We show that our tool-kit can seamlessly trigger neural activity waves across multiple layers in the network, followed by simultaneous self-organization of inter-layer weights, effectively speeding up the process of self-organization. We also demonstrate that the algorithm allows us to self-organize a wide variety of feedforward neural architectures, like multi-layer retinotopic layers and autoencoders. The ability to self-organize large networks of spiking units in a modular fashion is extremely relevant for the field of neuromorphic computing. Additionally, the framework established will be very useful for self-organizing large-scale models of the brain.

\section{Related Work}

Modeling the self-organization of neural networks (NNs) dates back many years, with the first demonstration being Fukushima's neocognitron \cite{fukushima1988neocognitron, fukushima1991handwritten}. It was built out of simple McCulloch-Pitts neuron units \cite{chakraverty2019mcculloch}, arranged in a hierarchical multi-layer neural network, capable of learning to perform pattern-recognition. Although the weights connecting the different layers were modified via unsupervised learning paradigms, the architecture of the network was hard-coded, which was inspired by Hubel and Wiesels' \cite{hubel1963shape} model of simple and complex cells in the visual cortex. The neocognitron design inspired modern day artificial NNs (ANNs) and convolutional NNs (CNNs) \cite{lecun1990handwritten}. ANNs and CNNs trained via global learning rules, like backpropagation, have been extremely successful in performing image-based tasks \cite{goodfellow2014generative, krizhevsky2012imagenet, amodei2016deep, silver2016mastering}. However, ANNs rely on hand-designed architectures for their functioning and suffer from the bottleneck of requiring massive datasets to learn efficiently. On the contrary, biological neural networks in the brain grow and self-organize a neural architecture that can generalize very well to innumerable datasets without requiring a massive training dataset. Inspired by the prowess of biological brains, the 3rd generation of NNs, namely SNNs \cite{ponulak2011introduction, maass1997networks, gruning2014spiking}, was proposed. SNNs are built out of `neuron' units that mirror the dynamics of living neurons. Although very promising, simulating large SNNs on conventional CPU's is very inefficient and time-consuming. The introduction of neuromorphic hardware, like IBM's TrueNorth \cite{merolla2014million} and Intel's Loihi \cite{davies2018loihi}, provided the right platform for simulating large (deep) SNNs for long time-periods, enabling networks to make inferences on a wide range of tasks. However, as SNNs are built out of dynamical units (spiking `neurons'), they are extremely sensitive to the initial wiring architecture. To overcome this challenge, authors in \cite{raghavan2019neural} have demonstrated an efficient self-organization routine to autonomously wire a two layered spiking neural network. The self-organization is driven by traveling spatio-temporal activity waves in the first layer, that ultimately lead to the formation of pooling structures. However, the strategy needs extensions for the self-organization of (deep) SNNs with multiple layers. The significant challenge in constructing multi-layer SNNs has been the decreasing spiking input signal intensities, which occur as a result of propagating through a layer, its weights and due to the mathematical nature of competition rules; ultimately making it extremely challenging for a signal instance to cause spikes in later layers \cite{meng2020spiking}. We overcome this challenge by proposing a dynamical framework that endows waves in the preceding layers with the ability to trigger input signals that initiate autonomous waves in subsequent layers. Triggering activity waves in subsequent layers (instead of independent, individual spikes) allows the network to establish an organized firing pattern throughout the network, in essence amplifying the signal received from the lower layers and passing information to higher layers without requiring additional transformation modules. 





\section{Modular SNN Tool-kit: Dynamical Systems Framework}

In order to build a scalable multi-layer SNN, we propose a dynamical systems framework for the self-organization algorithm. It utilizes the following key concepts of (i) emergent spatio-temporal waves of firing neurons, (ii) dynamic learning rules for updating inter-layer weights and (iii) non-linear activation and input$\addslash$output competition rules between layers to build a modular spiking sub-structure. The modular spiking sub-structure can be stacked to form multi-layered SNNs with an arbitrary number of layers that self-organize into a wide variety of connectivity architectures. The following sections describe the tool-kit required to build a single module that can be seamlessly stacked to self-organize multi-layer SNN architectures. We describe our framework by discussing the SNN model that generates waves and the learning$\addslash$competition rules that achieve inter-layer connectivity. 

\subsection{Governing Equations of "Neuronal Waves"}

The essential building block for SNNs is a spiking neuron model that describes the state of every single neuron over time, often represented by a dynamical system. Here, we choose a modified version of the popular Leaky-Integrate-and-Fire (LIF) model with an additional adjacency matrix term and input term (from preceding layers), coupled with a dynamical threshold equation. The vectorized governing equations for each layer $l$ reads
\vspace{4pt}
\begin{equation}\label{eq:LIF}
\begin{split}
    \frac{d}{dt}\bm{v} &= -\frac{1}{\tau_v}\bm{v} + \textbf{S}\mathcal{H}(\bm{v}-\bm{\theta}) + \textbf{S}^x \bm{x}\\
    \frac{d}{dt}\bm{\theta} &= -\frac{1}{\tau_\theta}(v^{th}-\bm{\theta}) \odot (1-\mathcal{H}(\bm{v}-\bm{\theta})) + \theta^+ \mathcal{H}(\bm{v}-\bm{\theta})\\
\end{split}
\end{equation}

where $v$ is the voltage, $\theta$ is the variable firing threshold, $\bm{x}$ is the input signal to this layer, $\mathcal{H}$ is the (element-wise) heavy-side function and $\odot$ denotes the Hadamard product. $\textbf{S}$ is the intra-layer adjacency matrix and $\textbf{S}^x$ is the spike input matrix. All vectors and matrices are elements of $\mathbb{R}^{n_l}$ and $\mathbb{R}^{n_l \times n_l}$ respectively, where $n_l$ is the number of neurons in layer $l$.
A neuron $i$ fires a spike when its voltage $v_i$ exceeds its threshold $\theta_i$. After firing, the neuron's voltage is reset to $v^{reset}$. The dynamic threshold equation for $\theta$ is governed by a homoeostasis mechanism to ensure that no neuron can spike excessively. Concretely, it increases $\theta$ by a rate $\theta^+$ whenever a neuron is spiking, until $\theta$ exceeds $v$ and the neuron fires no more. It then decays $\theta$ exponentially to a default threshold $v^{th}$. All additional hyper-parameters are summarized in the appendix.


$\textbf{S}\in \mathbb{R}^{n_l \times n_l}$ encodes the spatial-connectivity of neurons within the layer (that can have arbitrary geometry \cite{raghavan2019neural}) and is biologically inspired \cite{kutscher2004local, xiong2010cells}. Authors in \cite{laing2001stationary} have since used the intra-layer connectivity to prove the existence of spatio-temporal wave states in both 1D and 2D geometries of connected spiking neurons.
In our multi-layer SNN, $\textbf{S}\mathcal{H}(\bm{v}-\bm{\theta})$ serves as a back-coupling term, crucial for the development of coherent wave dynamics in subsequent layers. The optional spike-input matrix $\textbf{S}^x \in \mathbb{R}^{n_l \times n_l}$ can be used to further control the input received from preceding layers.
We encode the geometry of the layer and an isotropic kernel with a tunable excitation and inhibition radius and amplitude factors into $\bf{S}$. The kernel leads to positive intra-layer neuronal connectivity inside the excitation radius $r^i$ and decaying negative connections outside the inhibition radius $r^o$. Concretely, the adjacency matrix with kernel is given by

\begin{equation}
    S_{i,j} =     
    \begin{cases}
    \begin{aligned}
        &a^i D_{i,j},       &     D_{i,j} &<r^i \\
        -&a^o e^{(-D_{i,j}/10)}, & D_{i,j} &>r^o 
    \end{aligned}
    \end{cases} 
\end{equation}

where $D_{i,j} \in \mathbb{R}^{n_l \times n_l}$ is the matrix of spatial distances between each neuron and $a^i$$\addslash$$a^o$ are the excitation and inhibition amplitude factors. One can now vary the kernel radii and other hyper-parameters to control the emergent wave properties and obtain an array of wave phenomena with interesting shapes and dynamics. A few exemplary wave regimes are depicted in figure-\ref{fig:flexibleFeatures}(B). 

\subsection{Learning Rules}

Having constructed a spontaneous spatio-temporal wave generator across multiple layers in the previous section, we implement a local STDP learning rule to update inter-layer connectivity based on the patterns of the emergent waves, in order to self-organize SNNs into a wide variety of architectures. 
STDP potentiates connections between neurons that spike within a short interval to each other and provides lower updates for those neurons that have distant spike-times. As a simple STDP rule, we use the Hebbian rule to only link the synchronous pre- and post-synaptic firings of neurons for the dynamic update of weights between the two connected layers. We note that there are many types of sophisticated STDP rules such as additive STDP or triplet STDP \cite{markram2011history,bichler2011unsupervised}, however, we use a rather simple rule to only emphasize the effectiveness of our contribution. The learning rule can be integrated into our dynamical system as the dynamical matrix equation:

\begin{equation}\label{eq:learning_rule}
    \frac{d}{dt}\textbf{W}^{(l_1)} = \eta (\bm{y}^{(l_1)} \otimes \bm{y}^{(l_2)})
\end{equation}

where $\eta$ is the learning rate, $\bm{y}^{(l_1)} \in \mathbb{R}^{n_{l_1}}$ and $\bm{y}^{(l_2)} \in \mathbb{R}^{n_{l_2}}$ denote the spiking output signals of the two layers that $\textbf{W}^{(l_1)}  \in \mathbb{R}^{n_{l_1} \times n_{l_2}}$ connects and $\otimes$ is the outer product of the two vectors. 
The specific variables coupled in eq-\ref{eq:learning_rule} can be customized to achieve various desired connectivity architectures. 

\subsection{Competition Rules}

In addition to the learning rules, we can also introduce various "competition rules" on the layer inputs and outputs to further localize connections with different strengths, to form pooling architectures.
For instance, by coupling the spiking outputs in eq-\ref{eq:learning_rule} with $\bm{y}^{(l_2)}$ filtered by a "winner-take-all" competition rule, one can enforce the formation of pools from $l_1$ to the maximum spiking neuron in $l_2$. An input spike signal $\bm{x}$ can similarly be filtered. The winner-take-all competition rule for a vector $\bm{x}$ reads:

\begin{equation}\label{eq:competition_rule}
    f^\mathcal{C}(\bm{x}): 
    \begin{cases}
        x_i = 0, \hspace{30pt}\forall x_i<\text{max}(x)  \\
        x_i = \text{max}(x), \hspace{5pt} \mathrm{otherwise.}
    \end{cases} 
\end{equation}

The competition rule $f^C$ works on each neuron $i$ within a layer $l$. From eq-\ref{eq:competition_rule}, many variations like "$k$-best-performers" and other competition rules can be derived and applied to achieve pools of different shapes and weightings throughout the layers. 

\subsection{Multi-layer SNN Learning Algorithm}

With the three building blocks (eq-\ref{eq:LIF}, eq-\ref{eq:learning_rule}, eq-\ref{eq:competition_rule}) established, the algorithmic flow of an input signal $\bm{x}^{(1)}$ of a layer ($l_1=1$) to the input $\bm{x}^{(2)}$ of the next layer ($l_2=2$) is elaborated in algorithm-\ref{alg:multiLayerSNN}. In algorithm-\ref{alg:multiLayerSNN}, LIF($\cdot$) stands shorthand for a time-integration pass through eq-\ref{eq:LIF} and $\mathcal{H}^{(1)}_{v,\theta}$ is the respective spike vector. Furthermore, $f^{\mathcal{C}^{(1)}_y}$ and $f^{\mathcal{C}^{(2)}_x}$ are the (optional) competition rules for the output of $l_1$ and input to $l_2$ respectively and $g(\cdot)$ denotes the activation function of the layer, which is a rectified linear unit (ReLU) in our case. As one can see, the entire algorithm is model-able as a large dynamical system – coupling the wave dynamics equations of individual layers with the weight dynamics equations given by STDP learning rules between the layers. We integrate all equations in time at the same time-level by using a Runge-Kutta-4 time-stepping scheme for numerical integration.

\begin{algorithm}[t]
	\DontPrintSemicolon
	\SetAlgoLined
	\caption{Multi-layer SNN Dynamical System}
	\label{alg:multiLayerSNN}
	\SetKwInOut{Input}{Input}\SetKwInOut{Output}{Output}
	
	\Input{Signal $\bm{x}^{(1)}(t)$ as input to input layer $l=1$.}
	\vspace{1pt}
	\Output{Weights $\textbf{W}^{(l)}(t)$ \& spiking outputs $\bm{y}^{(l)}(t)$ for all layers $l\geq1$.}
	\vspace{5pt}
	\For{ t = $1...N_t$ in $\Delta t$ time-steps}
	{
		\For{ l = $1...N_l$ in layers}
		{
		\begin{math}
		\begin{aligned}
		\vspace{10pt}
		    &\mathcal{H}^{(l)}_{v,\theta} &&\mapsfrom \text{LIF}^{(l)}(\bm{x}^{(l)},\Delta t)
		    &&\textit{     integrate input with LIF by $\Delta t$ }\\
            &\bm{y}^{(l)} &&\mapsfrom f^{\mathcal{C}^{(l)}_y}(\mathcal{H}^{(l)}_{v,\theta}) 
		    &&\textit{     apply output competition rule to spikes}
		\end{aligned}
		\end{math}
		
        \vspace{5pt}
		\If{l$\geq$ 2}
		{
		\vspace{3pt}
		\textbf{W}$^{(l-1)} \mapsfrom $ \text{LR}$^{(l-1)}(\bm{y}^{(l-1)},\bm{y}^{(l)},\Delta t)$ \text{            integrate learning rule of preceding weights}
		\vspace{3pt}
		}
		\vspace{-1pt}
		\begin{math}
		\begin{aligned}
            &\bm{z}^{(l+1)} &&\mapsfrom \textbf{W}^{(l)} \bm{y}^{(l)} 
		    &&\text{     multiply local weights to output signal}\\
            &\bm{a}^{(l+1)} &&\mapsfrom g(\bm{z}^{(l+1)}) 
		    &&\text{     apply activation function}\\
            &\bm{x}^{(l+1)} &&\mapsfrom f^{\mathcal{C}^{(l+1)}_x}(\bm{a}^{(l+1)}) 
		    &&\text{     apply input competition rule to obtain signal for next layer}
		\end{aligned}
		\end{math}
        }
	}
\end{algorithm}

\section{Self-organizing Multi-layer Spiking Neural Networks}

\begin{figure}[t]
    \centering
    \includegraphics[scale=0.57]{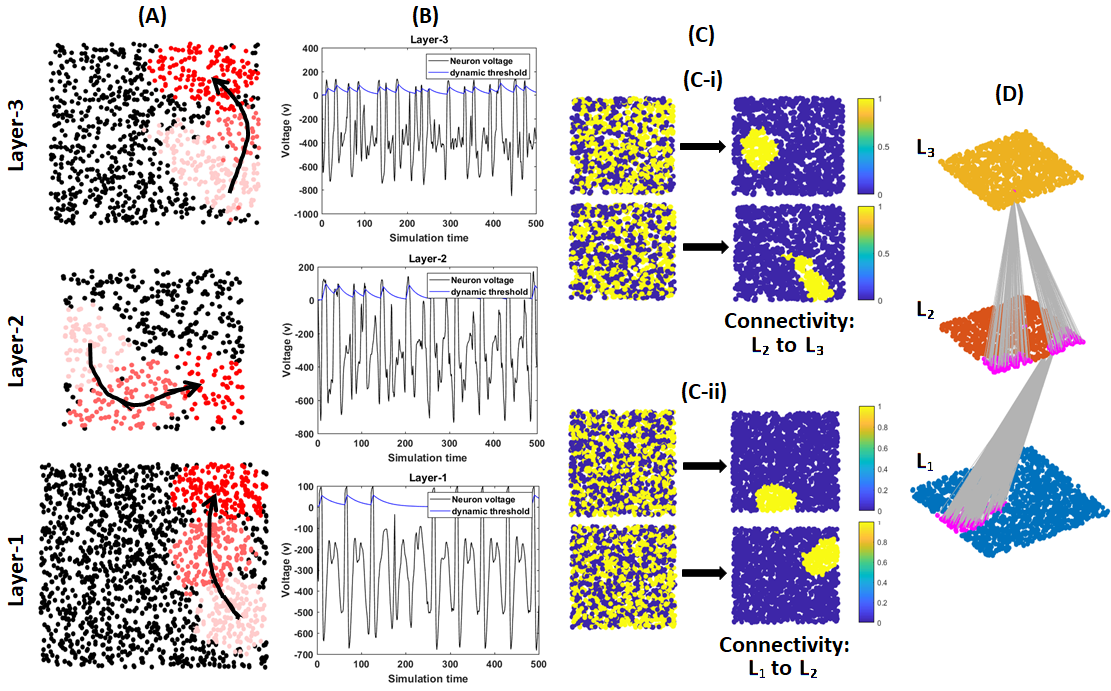}
    \caption{\textbf{Self-organizing multi-layer spiking neural networks} \textbf{(A)} Emergent spatio-temporal waves in $L_1$ trigger neuronal waves in higher layers ($L_2$, $L_3$). Black nodes indicate the neuron 
    positions within a layer and shades of red depict firing nodes. The lighter red represents nodes that fired at an earlier time-point. Lighter red to dark red captures the motion of the waves on each layer.
    \textbf{(B)} Tracking the voltage $v$ of a single neuron in each layer over time. The neuron `fires' when the $v$ crosses its dynamic threshold (blue line). 
    \textbf{(C)}  Self-organization process transforms a randomly wired inter-layer connectivity (left of the arrow) to a pooling architecture (right), wherein units in higher layers ($L_2$, $L_3$) are connected to a spatial patch of nodes in its preceding layer. Each subplot displays the connectivity of a single unit in a higher layer to all units in the preceding layer. Yellow$\addslash$blue represent regions with$\addslash$without presence of connections.
    Connectivity of 4 units each in $L_3$ and $L_2$ are depicted in C-i and C-ii respectively. 
    \textbf{(D)} 3D rendering of the final self-organized architecture. 
    }
    \label{fig:selfOrg_neuralArch}
\end{figure}

The modular tool-kit introduced in the previous section enables the efficient, autonomous self-organization of large multi-layer SNNs. The key ingredients required for self-organization 
are (i) traveling 
waves that emerge simultaneously across multiple layers and (ii) a dynamic learning rule that 
tunes the connectivity between any two layers based on the properties of the waves tiling the layers. We demonstrate the entire self-organization process in figure-\ref{fig:selfOrg_neuralArch} (moving from left to right). The two major components of the self-organization process are elaborated in the following subsections. 

\subsection{Emergent activity waves in multiple layers}

Stochastic communication between spiking neurons in layer-1 arranged in a local-excitation, global inhibition connectivity leads to the emergence of spontaneous traveling activity waves within the layer. 
The 
waves in layer-1 trigger waves in layer-2 that subsequently initiates 
waves in layer-3. 
The traveling waves across the 3 layers are depicted in figure-\ref{fig:selfOrg_neuralArch}A. We observe that the 
algorithm enables the motion of waves in higher layers without the need for a constant stimulation from the lower layers. In other words, the wave activity in higher layers, once triggered, can `stay alive' even if there is no spiking activity in the lower layers. Another key property of the traveling waves in the higher layers is that they have their own autonomy$\addslash$`curiosity' to explore different regions within the layer. The level of `curiosity' is dependent on the input from the preceeding layer and the strength of intra-layer connectivity. This forces the wave to not arbitrarily stray away from the source of the input-signal.

We also point out that 
waves in any layer are observed primarily due to the spiking dynamics of individual neurons. In figure-\ref{fig:selfOrg_neuralArch}B, we show the voltage trace of one neuron within each layer along with its spiking threshold. A neuron fires only when its voltage surpasses the spiking threshold, and the spiking frequency within each layer governs the dynamics of the activity wave.

\subsection{Local learning rules leads to self-organization}

The activity waves generated in each layer serve as a signal to modify their inter-layer weights. Along with the `signal', we need local learning rules to update inter-layer connections. Here, we use  Hebbian-based STDP rules (described in section 3.2) coupled with competition rules (described in section 3.3) to update inter-layer weights. In figure-2C, we depict the simultaneous activity-wave driven self-organization across multiple layers. The connectivity between the layers go from a random configuration to pooling structures between the layers, guided by the dynamics of the activity wave. A final self-organized multi-layer spiking network is rendered in 3D in figure-2D.

\section{Flexibility enabled by the dynamical systems framework}

The framework established in the previous section is the first demonstration of autonomous self-organization of a multi-layer spiking network, without the need for any additional transformation modules to connect subsequent layers. 

\begin{figure}[t]
    \centering
    \includegraphics[scale=0.48]{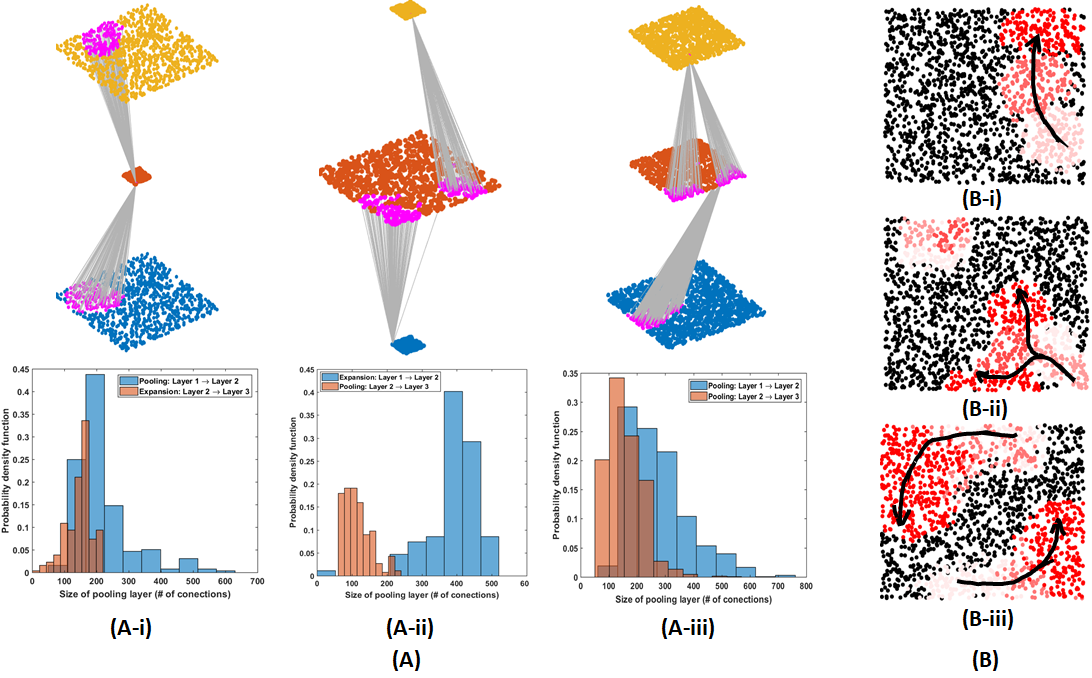}
    \caption{\textbf{Flexibility of the framework}. \textbf{(A)} \textit{Self-organizing a variety of neural architectures}: (A-i) Pooling followed by expansion (autoencoder) (A-ii) Expansion followed by pooling (A-iii) Consecutive pooling structures. Histograms capture the sizes of emergent pooling and expansion structures in the self-organized network. \textbf{(B)} \textit{Regimes of wave dynamics}: (B-i) Stable single wave (B-ii) Unstable splitting and merging waves (B-iii) Stable periodically rotating fluid-like wave.}
    \label{fig:flexibleFeatures}
\end{figure}

In this section, we demonstrate that designing the modular tool-kit in a dynamical systems framework naturally endows our system with flexible features. The modular construction of different layers allows us to tune the emergent wave dynamics on each layer, ultimately resulting in different self-organized architectures. The wave dynamics in each layer can be tuned by varying (i) excitation$\addslash$inhibition connectivity ($r^i$, $r^o$) between neurons within every layer and (ii) by altering the time-constants and other hyper-parameters governing the spiking dynamics of neurons in each layer. In figure-\ref{fig:flexibleFeatures}B, we portray a broad range of wave dynamics achievable on the layers of the network. 

Along with varying wave dynamics, modifying the size and shape of waves across different layers, and the number of nodes in each layer, we are able to self-organize a wide variety of multi-layer NN architectures (figure-\ref{fig:flexibleFeatures}). Here, we demonstrate efficient self-organization of three common neural architectures: (i) (Self-organized autoencoder) Pooling followed by expansion  
, (ii) Expansion followed by a pooling layer 
, (iii) Consecutive pooling operations (Self-organized retinotopic pooling structure) 
. The histograms in figure-\ref{fig:flexibleFeatures} capture the size of the self-organized pooling and expansion structures between the layers. The size of a pooling structure from $L_1$ $\rightarrow$ $L_2$ is the number of connections a single node in $L_2$ makes with nodes in $L_1$, while the size of the expansion structure from $L_2$ $\rightarrow$ $L_3$ is the number of connections a single node in $L_2$ makes with nodes in $L_3$. As the pooling and expansion structures follow a sharp uni-modal distribution, we infer that our algorithm imposes a tight control over the size of the self-organized structures. 


\section{Functionality: Real-time Unsupervised Feature Extraction}

In the previous section, we have demonstrated that spiking networks can be self-organized into a wide variety of architectures. In this section, we show that these networks are functional. In a preliminary assessment of semi-supervised classification on MNIST, we solely train a linear classifier appended to the end of an SNN self-organized by noise (without modifying SNN weights by back-propagation). The train$\addslash$test accuracy was consistent across multiple 3-layered SNNs averaging at 96.5$\%$$\addslash$93$\%$.

\begin{figure}[t]
    \centering
    \includegraphics[scale=0.62]{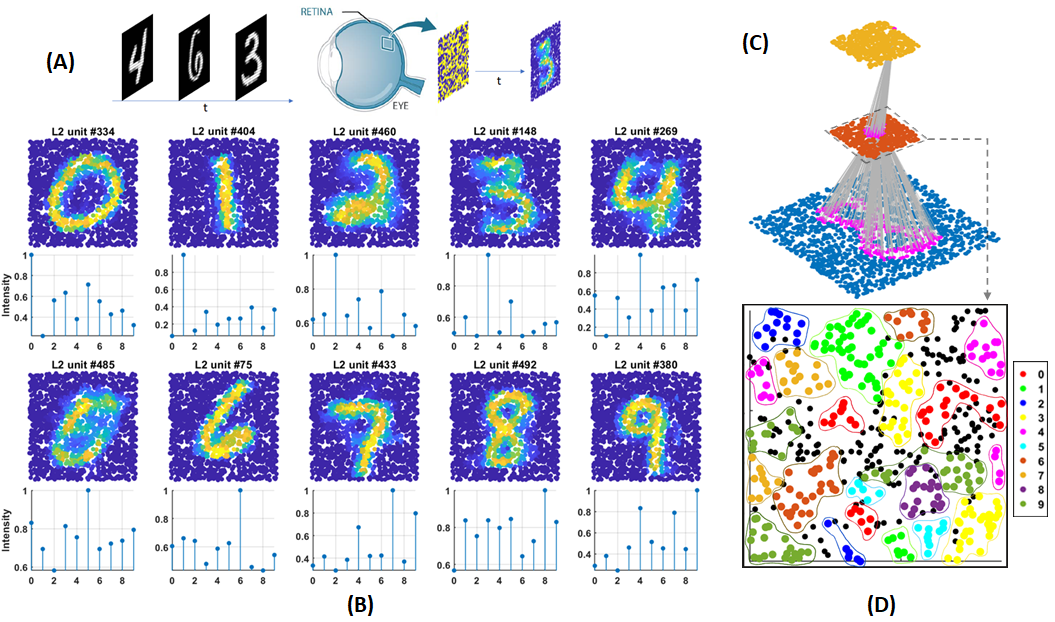}
    \caption{\textbf{Unsupervised learning of self-organized networks}. \textbf{(A)} Schematic of bio-inspired real-time learning: a 3-layered SNN learns on 2000 images, while being forward-integrated in time; it tests on circa 8000 images. \textbf{(B)} Unsupervised feature extraction forms pools that resemble MNIST digits: $\textbf{W}^{(1)}$ weights of 10 exemplary $L_2$ neurons connecting to displayed $L_1$ neurons that form pools in shapes of digits. The respective tuning curves of each $L_2$ unit shows the (0-1-scaled) mean output spike intensities to input spikes of all kinds of digits in the test set $-$ demonstrating the specialized $L_2$ unit spiking most intensely for one specific digit. \textbf{(C)} Exemplary connectivity pattern of the 3-layered network: pooling connection in shape of an `8'. \textbf{(D)} Coherent learning clusters in the $L_2$ that each, as a local group, specialize on learning$\addslash$classifying a certain class of input digit.}
    \label{fig:unsupervisedLearn}
\end{figure}

For the task of unsupervised feature extraction, we feed a stream of images as input to the algorithm in real-time, with a frame rate of one image every 5 seconds, while time-integrating the multi-layered SNN (figure-\ref{fig:unsupervisedLearn}). 
As a structured image-input is available, the parameter regime for the input layer ($L_1$) is chosen to ensure that noisy clusters of firing neurons shaped like the input image (here, MNIST digits) with spatio-temporal oscillations appear. Although there are no activity waves in $L_1$, we demonstrate that waves \textbf{will} still emerge in the subsequent layers. 

The local learning rules coupled with competition rules enable many $L_2$ neurons to extract features from the input image (MNIST digits). Also, certain $L_2$ units \textbf{specialize} on a single class of MNIST digits. The specialization of $L_2$ units for a single class of MNIST digits is clearly observed by visualizing its self-organized connectivity to the input-layer and its tuning curves, both depicted in figure-\ref{fig:unsupervisedLearn}B. The tuning curve for an $L_2$ unit is generated by feeding 10 classes of MNIST digits to the network and recording its spiking intensity. For instance, in figure-\ref{fig:unsupervisedLearn}B, $L_2$ unit $\#$404 has a connectivity to the input-layer that resembles MNIST digit `1' and its tuning curve (plotted below) confirms that $L_2$ unit $\#$404 maximally spikes when MNIST digits of class '1' are fed as input. Another interesting feature of our self-organization algorithm is that the neurons in $L_2$ that specialize for certain classes of MNIST digits, also \textbf{spatially cluster} within the layer. The spatial clustering of $L_2$ units for different MNIST classes are shown in figure-\ref{fig:unsupervisedLearn}D. The different node-colors correspond to neurons in $L_2$ that  specialize to different MNIST classes.  The spatial clustering of input-classes in $L_2$ is a direct consequence of the emergent spatio-temporal waves in $L_2$. Since the inter-layer connectivity is randomly initialized (mean: $\mu=1$, std. dev. $\sigma=0.5$) at $t=0$, even if a learning rule enabled the learning and increased specialization of certain $L_2$ units, one would not observe the formation of any type of spatial clustering of input-classes, i.e. the distribution of specialized neurons would be arbitrary, if it was not for the wave. The spatio-temporal wave in $L_2$ enables the formation of spatially coherent connections that proceed to become specialized coherent learning structures within $L_2$.

\section{Discussion}
In this paper, we address an important question of how large artificial computational machines could build and organize themselves autonomously without any involved human intervention. Currently, architectures of artificial systems are obtained after hours of painstaking hand parameter tuning. Inspired by the growth and self-organization of complex architectures in the brain, we introduce a dynamical systems framework to utilize emergent spatio-temporal activity waves to autonomously self-organize a multi-layer spiking neural network into a wide variety of architectures.

Our work has shed light on the importance of spatio-temporal neural computation. Most ANNs and their training algorithms do not take into account the spatial positions of their constituent `neurons' (computational units). Here, SNNs are built out of neurons with a distribution in 3D space relevant to the computation. The spatial relationship between constituent neurons is enforced by adjacency matrices, which leads to  biologically relevant phenomena like propagating neuronal activity waves and spatial clustering of units in higher layers that specialize for different classes of inputs. As emergent neuronal waves in the layer are a key biological phenomena, we believe that the AI$\addslash$ML community should consider spatial connectivity to build systems that are more `brain-like'.

The spatial clustering of functionality in the biological brain and the presence of spontaneous neuronal activity waves spanning the entire brain during development, suggests that our bio-inspired learning algorithm is an effective future direction for the development of computational neuroscience models and bio-inspired machine-learning tools. 


\section{Broader Impact}

AI has grown by leaps and bounds over the last decade and has become ubiquitous across a large number of industries. AI and neural networks have been implemented for real-time decision making in self-driving cars, have enabled data-driven diagnosis in hospitals and have enhanced the comforts at home by effectively being integrated into household appliances via IoT sensors. 

Although AI technology and neural networks are being actively incorporated in multiple industries to perform a wide range of tasks, discovering the right architecture for a particular task$\addslash$application continues to remain an ordeal. In scenarios, where effective neural network architectures have been discovered, they remain rigid to changes in input-size and might require a lot of pre-processing of the raw input before they can be fed to the network. Also, current methods for building neural networks are not suited for the flexible addition or removal of concurrent data streams. 

For example, mass produced camera technology that provides real-time data feeds from distributed cameras and drones deployed across the world can be simultaneously processed by neural networks to monitor climate change, agriculture, disaster prone regions and to assist policy makers and society planners to refine current practices. 

To do so, we need to construct neural networks that can simultaneously process multiple image data-streams and subsequently make intelligent decisions. Conventionally, neural network architectures are hand-designed to process concurrent feed from distributed cameras , based on the following parameters: (i) number of data-streams ($\#$ of input-cameras), (ii) data structure ($\#$ of pixels), (iii) the input frame-rate ($\#$ of images captured per second) to name a few. The current network architecture cannot autonomously adapt itself to the addition of new data-streams (new camera installations), or to updates in the data resolution, or to changes to the data-sampling rate. The lack of flexibility forces an engineer (or an AI resource provider) to constantly hand-tune and update their networks for inevitable changes to the camera-sensor network! 

In this paper, we propose a novel paradigm to wire large neural networks. Inspired by wiring of neural circuits in the growing brain of an infant, we can autonomously self-organize the connectivity of artificial neural networks. Wiring of networks via self-organization endows networks with the additional flexibility to quickly adapt to changes in the input `structure', changes in the number of input data-streams, eliminating the requirement of human intervention!

Also, as our algorithm is well-suited for networks built out of spiking units, we can directly implement flexible self-organization of networks on neuromorphic hardware. Neuromorphic hardware has recently gained a lot of traction for their low-power consumption, reduced latency and their on-chip learning functionality (unlike edge devices that can only perform inference).

\printbibliography

\end{document}